\documentclass[letterpaper]{article} 
\usepackage{aaai25}  
\usepackage{times}  
\usepackage{helvet}  
\usepackage{courier}  
\usepackage[hyphens]{url}  
\usepackage{graphicx} 
\usepackage{natbib}  
\usepackage{caption} 
\usepackage{algorithm}
\usepackage{algorithmic}

\usepackage{amsmath}
\usepackage{url}
\usepackage{multirow}
\usepackage{graphicx}
\usepackage{booktabs}
\usepackage{amssymb}

\usepackage{newfloat}
\usepackage{listings}
\DeclareCaptionStyle{ruled}{labelfont=normalfont,labelsep=colon,strut=off} 
\floatstyle{ruled}
\newfloat{listing}{tb}{lst}{}
\floatname{listing}{Listing}
\title{Supportiveness-based Knowledge Rewriting for Retrieval-augmented \\ Language Modeling}
\author {
    Zile Qiao\textsuperscript{\rm 1},
    Wei Ye\textsuperscript{\rm 2},
    Yong Jiang\textsuperscript{\rm 3}
    Tong Mo\textsuperscript{\rm 1}
    Pengjun Xie\textsuperscript{\rm 3}
    Weiping Li\textsuperscript{\rm 1}
    Fei Huang\textsuperscript{\rm 3}
    Shikun Zhang\textsuperscript{\rm 2}
}
\affiliations {
    \textsuperscript{\rm 1}School of Software and Microelectronics, Peking University\\
    \textsuperscript{\rm 2}National Engineering Research Center for Software Engineering, Peking University\\
    \textsuperscript{\rm 3}Alibaba Group\\
    \{zileq, wye, zhangsk\}@pku.edu.cn, \{motong, wpli\}@ss.pku.edu.cn, \{yongjiang.jy, chengchen.xpj, f.huang\}@alibaba-inc.com
}

\usepackage{bibentry}

\begin{document}

\maketitle

\begin{abstract}
Retrieval-augmented language models (RALMs) have recently shown great potential in mitigating the limitations of implicit knowledge in LLMs, such as untimely updating of the latest expertise and unreliable retention of long-tail knowledge. However, since the external knowledge base, as well as the retriever, can not guarantee reliability,  potentially leading to the knowledge retrieved not being helpful or even misleading for LLM generation. In this paper, we introduce Supportiveness-based Knowledge Rewriting (SKR), a robust and pluggable knowledge rewriter inherently optimized for LLM generation. Specifically, we introduce the novel concept of "supportiveness"—which represents how effectively a knowledge piece facilitates downstream tasks.
Based on supportiveness, we first design a training data curation strategy for our rewriter model, effectively identifying and filtering out poor or irrelevant rewrites (e.g., with low supportiveness scores) to improve data efficacy. We then introduce the direct preference optimization (DPO) algorithm to align the generated rewrites to optimal supportiveness, guiding the rewriter model to summarize augmented content that better improves the final response. Comprehensive evaluations across six popular knowledge-intensive tasks and four LLMs have demonstrated the effectiveness and superiority of SKR. With only 7B parameters, SKR has shown better knowledge rewriting capability over GPT-4, the current state-of-the-art general-purpose LLM.
\end{abstract}

\section{Introduction}

Recent advances in large language models (LLMs)\cite{OpenAI2023GPT4TR, Wang2022SelfInstructAL, Touvron2023LLaMAOA, anil2023palm} have significantly enhanced their performance in various natural language processing tasks. Pre-training on large-scale unsupervised corpora enables LLMs to store extensive knowledge within their parameters. However, updating LLMs with the most recent information is challenging due to the high cost of training\cite{Lin2023RADITRD}. Additionally, valuable but sensitive information is often excluded from LLM training to prevent data leakage~\cite{Huang2022AreLP, Carlini2020ExtractingTD}. Consistently and reliably retaining long-tail knowledge remains also a bottleneck~\cite{Wang2023ShallWP}. Retrieval-Augmented Language Modeling (RALM) addresses these issues by integrating retrieved non-parametric knowledge with LLMs~\cite{Borgeaud2021ImprovingLM, Lin2023RADITRD, Shi2023REPLUGRB, Ram2023InContextRL}. By leveraging external sources, RALM has achieved remarkable results in various language-based tasks.

\begin{figure}[t]
\centering
\includegraphics[width=0.9\linewidth]{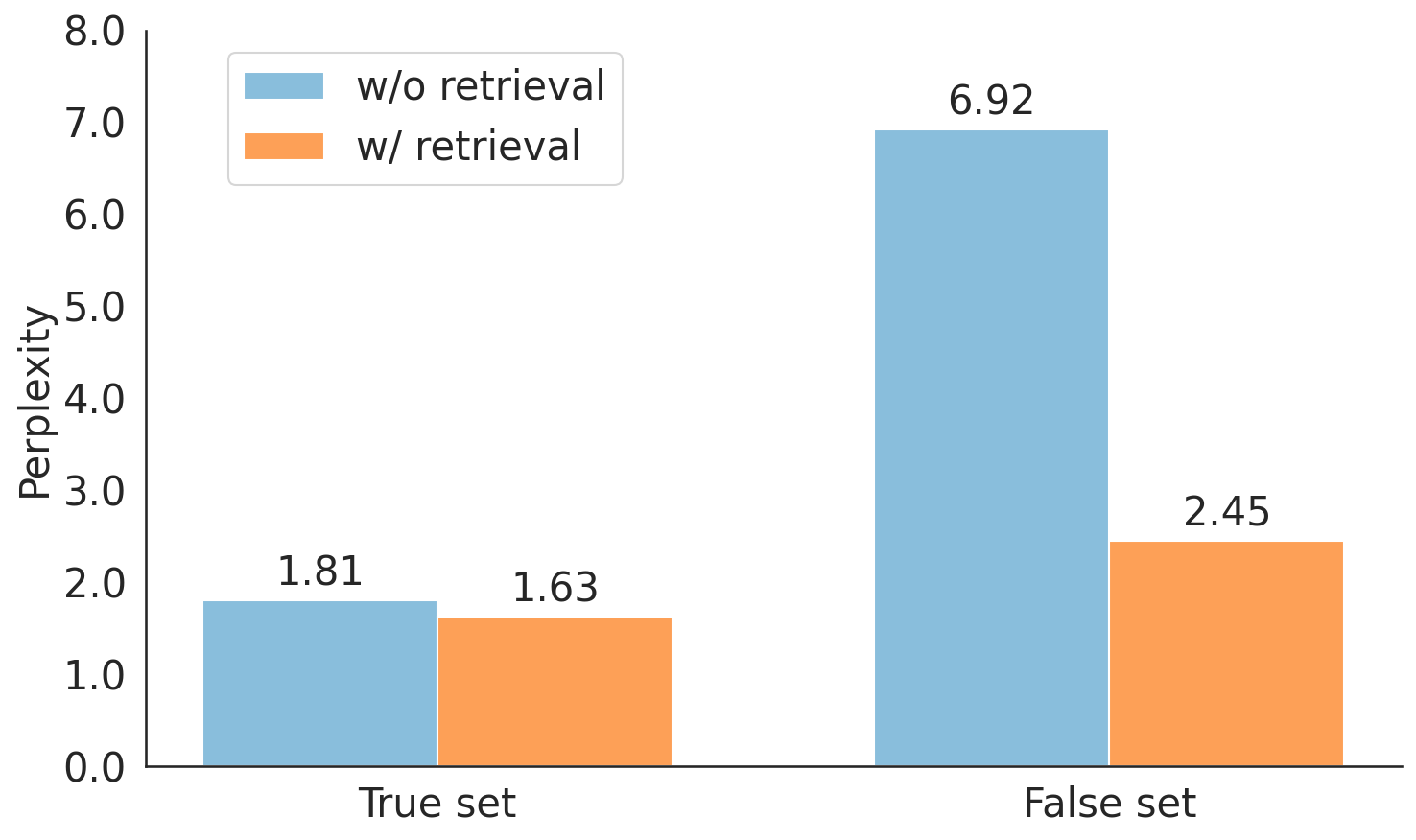}
\caption{To measure the change in perplexity induced by correct knowledge, we sampled 500 correctly answered questions (True set) and 500 incorrectly answered questions (False set) from TriviaQA. We then calculate the perplexity before and after introducing retrieved knowledge, where the retrieved knowledge included the correct answers.}
\label{fig:0}
\end{figure}

However, the significant noise and misleading information in the retrieved content can impair the performance of RALM. Recent efforts have aimed to mitigate this issue by rewriting the retrieved content. 
CoN~\cite{yu2023chainofnote} and RECOMP~\cite{Xu2023RECOMPIR} leverage off-the-shelf strong LLMs as supervision signal providers to train the rewrite. 
More recently, \citet{Wang2023LearningTF, jiang2024longllmlinguaacceleratingenhancingllms} and \citet{jin2024biderbridgingknowledgeinconsistency} have attempted to measure the quality of retrieved knowledge by assessing the perplexity (or probability) changes in the outputs of LLMs induced by retrieved knowledge, providing supervision signals for model training. However, these methods overlook the influence of the internal knowledge of LLMs. 
When an LLM's internal knowledge is sufficient to accurately respond to a particular query, the incorporation of retrieved knowledge, even if entirely accurate, may not significantly alter the output probabilities of the LLM, as illustrated in Figure 1. This phenomenon leads to feedback from the LLM that does not accurately reflect the quality of the retrieved knowledge.


To address this issue, we introduce the concept of "supportiveness". 
Beyond analyzing the variations in LLM output perplexity induced by retrieval knowledge, we further examined how well the internal knowledge of the LLM covers various queries. This approach enabled us to attain a more precise evaluation of the quality of the retrieved knowledge.
Building on supportiveness, we present Supportive Knowledge Rewriting (SKR), an abstractive rewriter inherently optimized for LLM generation. This rewriter is trained using two supportiveness-based mechanisms.


The first one is Supportiveness-based Rewrite Data Generation. Though powerful general-purpose LLMs (e.g., GPT-4~\cite{openai2023gpt4}) are widely known for their capability of serving as data annotators~\cite{yu2023chainofnote, Wang2022SelfInstructAL}, directly using LLM-generated rewrite data to train the rewriter may yield sub-optimal performance due to the bias and potential inaccuracy in LLM responses. Therefore, we use supportiveness to assess the quality of automatically generated rewrite data and then filter and refine it. In particular, we initially utilize a powerful LLM to create multiple different rewrites for each query based on a wide array of downstream tasks and retrieval data, serving as our preliminary draft dataset. We then engage a white-box LLM for each rewritten text to determine its supportiveness. 
Based on the supportiveness score, we refined the draft dataset, and the refined datasets were subsequently used for supervised fine-tuning (SFT).

The second mechanism is Supportiveness-guided Alignment, a training strategy that aligns the generated rewrite to optimal supportiveness using Direct Preference Optimization (DPO)~\cite{Rafailov2023DirectPO}. Specifically, we pair two different rewrites for the same query and calculate the supportiveness difference within each pair. Pairs with more significant discrepancies are selected for DPO training. In each selected pair, the rewrite with higher supportiveness serves as the positive sample, while the one with lower supportiveness serves as the negative sample. This process encourages SKR to generate rewrites that better support LLMs in performing downstream tasks, 

Our contributions are as follows:
\begin{itemize}
    \item We conceptualized "supportiveness", offering a novel perspective to assess how effectively an augmented knowledge piece contributes to a specific query.
    \item Leveraging supportiveness, we devise an effective knowledge rewriter by incorporating (1) a supportiveness-data rewrite data generation method that improves data efficacy and (2) a direct preference training mechanism to align rewritten text to optimal supportiveness better. Experiments conducted on six datasets have thoroughly validated the effectiveness and generalization of SKR. With merely 7B parameters, SKR has exhibited a knowledge rewriting capability that surpasses that of GPT-4
    \item Further experimental analysis validated that SKR effectively removes noise and misleading information in retrieval data, achieving a compression rate exceeding 7x. These findings can inspire future RALM research by highlighting the potential of supportiveness exploitation and knowledge rewriting.
\end{itemize}

\begin{figure*}[th]
\centering
\includegraphics[width=1\linewidth]{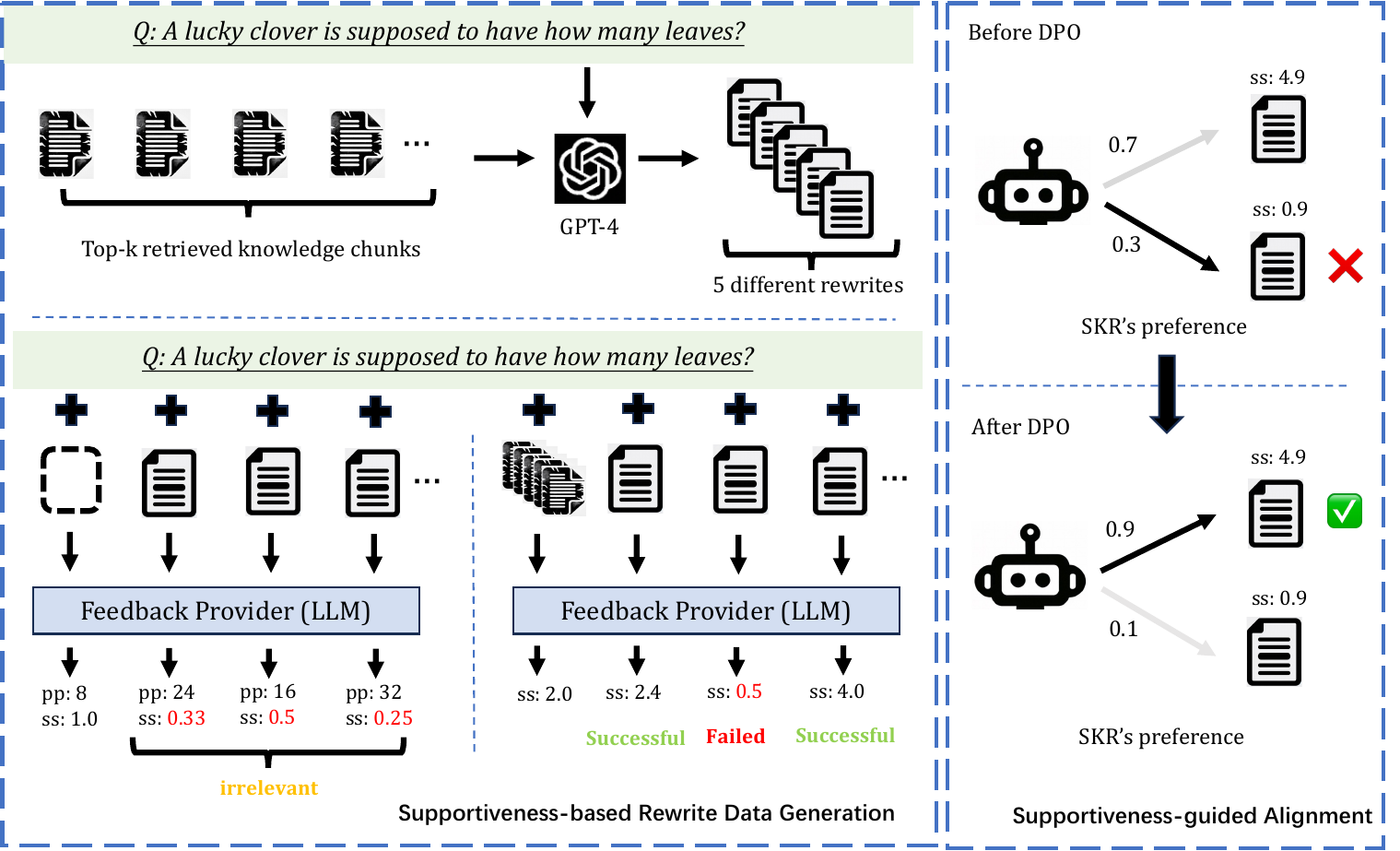}
\caption{A visual illustration of the proposed Supportiveness-based Rewrite Data Generation and Supportiveness-guided Alignment. ``pp'' represents perplexity, and ``ss'' denotes the supportiveness score.}
\label{fig:2}
\end{figure*}

\section{Method}

\subsection{Architecture}
\textbf{Retriever\quad}
Following~\cite{Lin2023RADITRD}, we utilize DRAGON+~\cite{lin2023train} as the retriever, which is a state-of-the-art dual-encoder model.
Given a query $q$ and a corpus $\mathcal{C}$, the query encoder $\mathbf{E}_q$ maps $q$ to an embedding $\mathbf{E}_q(q)$ and the document encoder $\mathbf{E}_d$ maps each text chunk $c \in \mathcal{C}$ to an embeddings $\mathbf{E}_d(c)$.
After that, we compute the similarity score $s(q,c)$ for each $c \in \mathcal{C}$ by dot product: $s(q,c) = \mathbf{E}_d(c) \cdot \mathbf{E}_q(q).$
The top-k relevant text chunks for $q$ are selected based on their similarity scores.

\noindent\textbf{Rewriter\quad}
To eliminate unhelpful or misleading information from retrieved knowledge, we proposed SKR as the knowledge rewriter. The rewriter is designed to refine retrieved knowledge through rewriting, aiming to optimize the utilization efficiency of external knowledge by LLMs.
Given a query $q$ and k candidate retrieval chunks \( c_q \in \mathcal{C}_q \), the input text $T$ to the rewriter is a concatenation of the query q and k retrieved knowledge chunks, separated by newline characters.
The rewriter generates rewrites in an autoregressive manner. 
The generated rewrite will serve as part of the prompt to assist the LLM in handling downstream tasks.

\noindent\textbf{In-Context Retrieval-Augmentation\quad}
For a given query $q$, 
retrieved knowledge chunks or the rewrites generated by the rewriter, we map them to a prompt $S$ according to the template to construct the input of language model to generate the response $y$ to the query $q$:
\begin{equation}
\begin{aligned}
& P_{LM}\left(y_1, y_2, \ldots, y_n\right)= \prod_{i=1}^n P_{LM}(y_i \mid S),
\end{aligned}
\end{equation}
where $y=\left\{y_1, y_2, \ldots, y_n\right\}$ denotes the sequence of tokens in the response.

\subsection{Supportiveness}
\label{sec:2.2.2}
Supportiveness indicates the degree to which a piece of knowledge assists in answering a specific query. To estimate supportiveness, we first need to employ an LLM (e.g. Mistral) to attempt to answer the current query twice, once with the integration of retrieved knowledge and once without it. Then, based on the correct label of the query, we calculate the perplexity of both outputs. 
The formal description of this process is as follows:
\begin{equation}
\begin{aligned}
& \mathcal{P}_{\text {raw }}= \exp \left(-\frac{1}{N} \sum_{i \in \text { target }} \log \left(P_{L M}\left(w_i \mid S_q \circ w_{<i}\right)\right)\right) \\
& \mathcal{P}_{\text {retrieval }}= \exp \left(-\frac{1}{N} \sum_{i \in \text { target }} \log \left(P_{L M}\left(w_i \mid S_q^r \circ w_{<i}\right)\right)\right),
\end{aligned}
\end{equation}
where $\mathcal{P}_{\text {retrieval}}$ and $\mathcal{P}_{\text {raw}}$ denote the perplexity of the output with retrieval and the output without retrieval, respectively. $\left\{w_1, \ldots, w_n\right\}$ denotes the target sequence of the query and $\circ$ denote the text concatenation.
For cases with multiple target sequences, we calculate the average of the perplexities computed from the numerous target sequences.
$S_q^r$ and $S_q$ represent the prompt with and without retrieval, which are shown in Appendix A. 

Consequently, we calculate the ratio of $\mathcal{P}_{\text{raw}}$ to $\mathcal{P}_{\text{retrieval}}$, denoted as $r(q, c)$, where \( c \) represents a specific knowledge piece.
Intuitively, the ratio $r(q, c)$ can represent the degree to which a specific knowledge piece supports the query. However, we note that when the LLM can easily answer the question without any retrieval knowledge (e.g., $\mathcal{P}_{\text{raw}}$ is relatively low), $r(q, c)$ cannot accurately measure the supportiveness of the knowledge piece $c$. In such cases, even if the knowledge piece $c$ contains key information to answer the query, it will not be reflected in the metric because LLM already possesses this knowledge.

To address this issue, we propose supportiveness score, denoted as $ss(q, c)$, which is calculated as:
\begin{equation}
\begin{aligned}
    & r(q,c) = {\mathcal{P}_\text{raw}}/{\mathcal{P}_\text{retrieval}} ,\\
    & ss(q, c) = r(q,c)/\sigma(\mathcal{P}_\text{raw}),
\end{aligned}
\end{equation}
where $\sigma$ denotes the sigmoid function. Compared to directly using the ratio of $\mathcal{P}_{\text{raw}}$ to $\mathcal{P}_{\text{retrieval}}$, the supportiveness score more accurately represents the supportiveness of a knowledge piece for a query, mitigating the bias introduced by LLM's internal knowledge.


\subsection{Supportiveness-based Rewrite Data Generation}
\label{sec:2.2}
In this section, we provide a detailed description of Supportiveness-guided Rewrite Data Generation (SRDG). 
The left part of Figure~\ref{fig:2} demonstrates a visual depiction of these approaches.

\subsubsection{Automated Rewrite Data Generation}
To train the rewriter, we first need to collect a substantial amount of rewrite data. Considering that manual annotation is resource-intensive and it is well-known that powerful general-purpose LLMs can serve as data annotators~\cite{yu2023chainofnote}, we employed the current state-of-the-art LLM, GPT-4, as a data annotator to generate our draft dataset. 
To ensure the diversity of the data, we used the training sets from a total of 10 datasets across three tasks (Open-Domain QA, Reading Comprehension, and Reasoning) as the sources for generating rewrite data. 
We used the template to prompt GPT-4 to generate rewrites, as shown in Appendix A.

To enhance the SKR's ability to generalize across varying lengths of context, we use the top-5 retrieved chunks for 80\% of the samples. For the remaining 20\% of the samples, we evenly distribute the use of the top-1, top-2, top-3, and top-4 retrieved knowledge chunks as prompts.
To further enrich the rewrite data and also to enable subsequent adjustments to the generation preference of SKR, we sampled five different rewrites for each query.

\subsubsection{Supportiveness-based Data Refinement}
Then, we introduce how to refine the draft dataset based on supportiveness. 
Through the supportiveness score, we are able to assess the degree to which a specific knowledge piece aids the query. 
Next, we will refine the draft dataset based on the supportiveness score. 
We categorize rewrites into three groups: ``irrelevant,'' ``failed,'' and ``successful''.

Given a query $q$, its retrieval data $c$, and the rewrites $\mathcal{R}= \left\{ r_1, r_2,  \ldots, r_n \right\}$ generated by GPT-4, we will first label the sample as "irrelevant" if the following conditions are met:
\begin{equation}
    \begin{aligned}
        & ss_{r_i} < 1 \quad  \text{for}\ r_i \in \mathcal{R}, \\
        & ss_{c} < 1,
    \end{aligned}
\end{equation}
where $ss_{r_i}$ and $ss_c$ denote the supportiveness score of $r_i$ and raw retrieval data, respectively. 
In this situation, we eliminate all of the rewritten content generated by GPT-4 and change the label rewrite of $c$ to the word "irrelevant". 
This scenario suggests that regardless of whether it is rewritten or not, the retrieved knowledge cannot assist the LLM in answering the query. 
Therefore, this refinement enhances the rewriter's ability to recognize unhelpful retrieved knowledge and further compresses the length of the rewrites to improve the overall efficiency of RALM.
Then, we will label the rewrite as ``failed'' if the following conditions are met: $ss_{r_i} / ss_c < \delta.$
This situation implies that the specific rewrite \( r_i \) is failed, as this rewrite, compared to the raw retrieval data, does not offer any additional assistance to the LLM. Therefore, we simply discard these rewrites.
The remaining samples are labeled as ``successful''. These, along with the 'irrelevant' samples, are then used for the training of SKR, employing a standard supervised fine-tuning approach:
\begin{equation}
\pi_{\mathrm{SFT}}=\max _\pi {\mathbb{E}} \log \pi(r \mid q \circ c),
\end{equation}
where \(\{ r_i \mid i = 1 \ldots k \}\) denotes \(k\) different rewrites for a given pair of $(q,c)$ and $\circ$ denotes text concatenation.
Through supportiveness-based data refinement, approximately 9 \% of the samples were labeled as "irrelevant" and modified, while 16\% were labeled as "failed" and discarded.

\subsection{Supportiveness-guided Alignment}
\label{sec:2.3}
After Supervised fine-tuning on the refined dataset, SKR is already capable of generating decent rewrites. 
To further strengthen SKR and make its outputs more supportive of LLMs in completing downstream tasks, we will alter the generative preferences of SKR to make it more inclined towards generating more supportive rewrites.

First, we construct preference data based on the draft dataset and the supportiveness score calculated for each rewrite.
Given a query $q$, the raw retrieval data $c$ and the rewrites set $\mathcal{R} = \left\{ r_1, r_2,  \ldots, r_n \right\}$ from draft dataset (we additionally include ``irrelevant'' in \(\mathcal{R}\) as an extra version of the rewrite), we construct the preference data $\mathcal{D}_p$ as follows:
\begin{equation}
    \mathcal{D}_p = \left\{ (x, r_w, r_l) \mid r_w, r_l \in \mathcal{R}, ss_w - ss_l > 1 \right\},
\end{equation}
where \(ss_w\) and \(ss_l\) respectively represent the supportiveness scores of the corresponding rewrites. $x = q \circ c$ denotes the input of SKR.
This construction implies that the supportiveness score of \(r_l\) in the preference data is significantly lower than that of \(r_w\).

Then, we employed DPO~\cite{Rafailov2023DirectPO} to align the generative preferences of SKR with the supportiveness of the rewrites. 
The objective of DPO training for SKR $\pi_{\mathrm{DPO}}$ can be written as,
\begin{equation}
    \begin{aligned}
    \pi_{\mathrm{DPO}} &= \max_\pi \underset{(x, r_w, r_l) \sim \mathcal{D}_p}{\mathbb{E}} \Bigg[ \log \sigma \Bigg( \beta \log \frac{\pi(r_w | x)}{\pi_{\mathrm{SFT}}(r_w | x)} \\
    &\quad - \beta \log \frac{\pi(r_l | x)}{\pi_{\mathrm{SFT}}(r_l | x)} \Bigg) \Bigg],
    \end{aligned}
\end{equation}
where $\beta$ is the hyperparameter for DPO training. The goal of this step is to refine the $\pi_{\mathrm{SFT}}$ by maximizing the likelihood of ranking the preferred $r_w$ over $r_l$.

\begin{table*}[ht]
\centering
\begin{tabular}{@{}lllllllllll@{}}
\toprule
\multirow{2}{*}{Model}       & \multirow{2}{*}{Rewriter} & \multirow{2}{*}{Retrieval} & \multirow{2}{*}{NQ} & \multirow{2}{*}{TQA} & \multirow{2}{*}{HoPo} & \multirow{2}{*}{MMLU} & \multirow{2}{*}{zsRE} & \multirow{2}{*}{WQ} & \multirow{2}{*}{Avg.} & \multirow{2}{*}{$\Delta_{EM}$} \\
                             &                           &                            &                     &                      &                       &                       &                       &                     &                       &                                \\ \midrule
\multirow{9}{*}{Llama-2-7B}  & -                         & -                          & 19.61               & 52.04                & 18.04                 & 39.96                 & 17.07                 & 19.01               & 27.62                 & -                              \\
                             & -                         & Top-1                      & 23.55               & 58.82                & 28.04                 & 43.17                 & 56.39                 & 19.59               & 38.26                 & -                              \\
                             & -                         & Top-5                      & 17.5                & 56.8                 & 25.43                 & 44.53                 & 42.4                  & 14.91               & 33.6                  & -                              \\ \cmidrule(l){2-11} 
                             & GPT-3.5                   & Top-5                      & 27.78               & 61.71                & 30.16                 & 45.99                 & 59.22                 & 21.3                & 41.03                 & 7.43                           \\
                             & GPT-4                     & Top-5                      & 29.82               & \textbf{63.76}       & 30.52                 & 47.77                 & 60.04                 & 22.83               & 42.46                 & 8.86                           \\
                             & RECOMP                    & Top-5                      & 27.02               & 61.04                & 28.29                 & -                     & -                     & -                   & -                     & -                              \\
                             & FILCO$\dagger$            & Top-5                      & 28.27               & 60.78                & 28.44                 & -                     & 59.27                 & 20.18               & -                     & -                              \\
                             & LongLLMLingua             & Top-5                      & 27.44               & 60.28                & 27.92                 & 44.23                 & 59.44                 & 21.01               & 40.05                 & -                              \\
                             & SKR-7B                    & Top-5                      & \textbf{31.42}      & 63.62                & \textbf{30.98}        & \textbf{48.45}        & \textbf{61.97}        & \textbf{25.24}      & \textbf{43.61}        & 10.01                          \\ \midrule
\multirow{9}{*}{Llama-2-70B} & -                         & -                          & 30.61               & 68.82                & 26.59                 & 62.84                 & 28.65                 & 22.21               & 39.95                 & -                              \\
                             & -                         & Top-1                      & 33.57               & 69.14                & 33.6                  & 65.58                 & 63.89                 & 25.01               & 48.47                 & -                              \\
                             & -                         & Top-5                      & 35.32               & 69.54                & 32.16                 & 66.15                 & 65.17                 & 25.17               & 48.92                 & -                              \\ \cmidrule(l){2-11} 
                             & GPT-3.5                   & Top-5                      & 35.68               & 69.47                & 36.24                 & 66.8                  & 65.91                 & 28.22               & 50.39                 & 1.47                           \\
                             & GPT-4                     & Top-5                      & 36.51               & 69.83                & 38.22                 & 67.09                 & \textbf{67.15}        & 30.15               & 51.49                 & 2.57                           \\
                             & RECOMP                    & Top-5                      & 35.47               & 69.44                & 34.47                 & -                     & -                     & -                   & -                     & -                              \\
                             & FILCO$\dagger$            & Top-5                      & 35.5                & 69.81                & 35.68                 & -                     & 66.12                 & 26.44               & -                     & -                              \\
                             & LongLLMLingua             & Top-5                      & 33.42               & 69.11                & 34.17                 & 65.78                 & 62.88                 & 28.78               & 49.02                 & -                              \\
                             & SKR-7B                    & Top-5                      & \textbf{37.97}      & \textbf{70.68}       & \textbf{38.24}        & \textbf{67.87}        & 66.91                 & \textbf{31.67}      & \textbf{52.22}        & 3.3                            \\ \bottomrule
\end{tabular}
\caption{\label{tab:main}
Main results.  All experimental results are presented using the EM (Exact Match) metrics,``$\Delta_{EM}$'' represents the average performance difference of the corresponding rewriting method compared to using ``Top-5'' retrieved data without any rewriting. ``$\dagger$'' denotes our implementations.}
\end{table*}

\section{Experiment Setup}
\label{sec:3}
\subsection{Implementation Details}
\label{sec:3.1}
\textbf{Automated Rewrite Data Generation\quad}
To construct the draft dataset, we utilized GPT-4 (gpt-4-1106) to automatically annotate rewrite data. We set the temperature to 1, "n" (the number of chat completion choices) to 10, and randomly sampled a certain number of responses to serve as the rewrite data. All other arguments were kept at their default settings.

\noindent\textbf{SKR\quad}
SKR was initialized with Mistral-7B, which also served as the feedback provider during supportiveness calculation. 
In both SFT and DPO, the hyperparameters used are consistent with those in zephyr~\cite{tunstall2023zephyr}, except for the learning rate, which we set to 1e-05 during the SFT process.
We used 10 datasets to train SKR, including four Open-domain QA datasets(CommensenseQA~\cite{Talmor2019CommonsenseQAAQ}, Yahoo! Answers, WikiQA~\cite{yang-etal-2015-wikiqa}, and FreebaseQA~\cite{jiang-etal-2019-freebaseqa}), four Reading comprehension tasks(COQA~\cite{Reddy2018CoQAAC}, DROP~\cite{dua-etal-2019-drop}, QuaRel~\cite{Tafjord2018QuaRelAD}, and SQuAD v2~\cite{rajpurkar-etal-2018-know}) and two reasoning datasets(GSM8K~\cite{Cobbe2021TrainingVT} and ECQA~\cite{Aggarwal2021ExplanationsFC}). Note that we did not use any evaluation datasets for training. 
Detailed dataset statistics can be found in Appendix B.

\noindent\textbf{Others\quad}
Most of our evaluations are conducted using vLLM~\cite{Kwon2023EfficientMM}. 
We conducted all training and evaluations on 4x NVIDIA A100 80GB GPUs. The process of SFT for training SKR took approximately 30 hours, and the DPO process took about 40 hours. The retrieval corpus used in this paper is Wikipedia dump 2018.
\subsection{Evaluation}
\textbf{Datasets\quad}
We have conducted evaluations on knowledge-intensive tasks and ensured that the dataset used for training the SKR does not appear in them. 
Specifically, we use Massive Multitask Language Understanding 
 (MMLU)~\cite{hendrycks2021measuring}, Natural Questions (NQ)~\cite{kwiatkowski-etal-2019-natural}, TriviaQA~\cite{joshi-etal-2017-triviaqa}, HotpotQA (HoPo)~\cite{Yang2018HotpotQAAD}, WebQuestions (WQ)~\cite{berant-etal-2013-semantic}, and zero-shot Relation Extraction~\cite{Levy2017ZeroShotRE} (zsRE) from KILT tasks~\cite{Petroni2020KILTAB}. We use dev split for Natural Questions, TriviaQA, and zsRE, and test split for MMLU, WebQuestions, and HotpotQA. All experimental results are presented using the EM (Exact Match) metrics. For the evaluations of these datasets, we use the same prompts as mentioned in Appendix A. 
 
\noindent\textbf{Baselines\quad}
We primarily compare SKR with GPT-3.5 and GPT-4~\cite{openai2023gpt4}, as well as two popular methods, Filco~\cite{Wang2023LearningTF} and RECOMP~\cite{Xu2023RECOMPIR}.
We use the same prompts for all the models.

\begin{table*}[t]
\centering
\begin{tabular}{@{}llllllll@{}}
\toprule
\multicolumn{1}{c}{\multirow{2}{*}{Models}} & \multicolumn{1}{c}{\multirow{2}{*}{Rewriter}} & \multicolumn{3}{c}{NQ}                 & \multicolumn{3}{c}{WQ}                 \\ \cmidrule(l){3-8} 
\multicolumn{1}{c}{}                        & \multicolumn{1}{c}{}                          & None  & Bad            & Random        & None  & Bad            & Random        \\ \midrule
\multirow{2}{*}{Llama-2-7B}                 & None                                          & 19.61 & 10.08          & 10.31         & 19.01 & 9.30           & 12.84         \\
                                            & SKR-7B                                        & -     & 20.11 (+10.03) & 19.55 (+9.24) & -     & 22.95 (+13.65) & 18.68 (+5.84) \\ \midrule
\multirow{2}{*}{Llama-2-13B}                & None                                          & 26.5  & 15.37          & 23.13         & 24.05 & 16.14          & 21.34         \\
                                            & SKR-7B                                        & -     & 24.91 (+9.54)  & 26.23(+3.10)  & -     & 27.77 (+11.63) & 24.26 (+2.92) \\ \midrule
\multirow{2}{*}{Mistral-7B}                 & None                                          & 25.73 & 18.25          & 19.78         & 21.8  & 15.80          & 19.13         \\
                                            & SKR-7B                                        & -     & 22.32 (+4.07)  & 24.19(+4.41)  & -     & 24.37 (+8.57)  & 21.22 (+2.09) \\ \bottomrule
\end{tabular}
\caption{\label{tab:noise}
Impact of noise and misleading information. ``Bad'' and ``Random'' denote using the Top-5 knowledge chunks respectively disturbed by the two methods described in section \ref{sec:misleading}, while ``None'' indicates not using any retrieved knowledge.}
\end{table*}

\section{Experiment Results}
\label{sec:4}
In this section, we report the experimental results (statistically significant with $p < 0.05$). Due to the page limit, we have placed the analysis results regarding the choice of retriever in Appendix D. Another analysis experiment related to the feedback provider is included in Appendix E.
\subsection{Main Results}
\label{sec:4.1}
We report the main results in Table~\ref{tab:main}. The experimental results using Llama-2-13B and Mistral-7B as the base model can be found in Appendix C. From these results, we make the following four observations:


\noindent\textbf{Negative Impact of Noise\quad}
Our first observation is that adding the Top-5 knowledge chunks, as opposed to just the Top-1 knowledge chunk, does not significantly enhance model performance as expected, even though the Top-5 knowledge chunks clearly contain more useful information (compared to the Top-1 setting, the Top-5 setting resulted in changes of -4.67, -2.48, +0.45, and -5.49 in the EM metric for Llama-2-7B, -13B, -70B, and Mistral-7B, respectively, on average across six different tasks). 
\textit{The additional noisy data introduced by more retrieved knowledge chunks has a significant negative impact on the LLMs, affirming the necessity of rewriting the retrieved knowledge.}

\noindent\textbf{Effectiveness of Supportiveness-based Rewrite\quad}
Our second observation is that \textit{the proposed SKR significantly enhances the LLMs' ability to utilize retrieved knowledge, markedly reducing the negative impact of noise on various LLMs} (compared to the Top-5 setting without a rewriter, SKR consistently yielded improvements of +10.01 and +3.30 in the EM metric for Llama-2-7B and Llama-2-70B, respectively, on average across six different tasks).
This result validates the capability of SKR to rewrite retrieved knowledge, confirming the effectiveness of this approach.

\noindent\textbf{Effectiveness of Supportiveness-based Training Strategy\quad}
Our third observation is that \textit{guided by the concept of supportiveness, SKR, with only 7B parameters, has exhibited a knowledge rewriting capability that surpasses representative powerful off-the-shelf LLMs, including the current state-of-the-art general-purpose LLM, GPT-4}. Compared to GPT-4, SKR yielded consistent improvements in the EM metric for Llama-2-7B and Llama-2-70B of +1.15 and +0.73, respectively, on average across six different tasks.
Compared to GPT-3.5, 
the improvements brought by SKR are even more pronounced.
These results demonstrate the effectiveness of our proposed supportiveness-based rewrite data
generation and supportiveness-based alignment methods. 
A detailed analysis of the individual contributions of these two methods will be presented in Ablation Studies.

\noindent\textbf{Performance Superiority Compared to Other Rewriting Methods\quad} The final observation is that \textit{SKR's performance significantly surpasses other rewriting methods.} SKR shows substantial performance gains across all six tasks compared to FILCO, RECOMP and LongLLMLingua. It's important to note that FILCO and RECOMP require training the rewriter on corresponding training sets, whereas SKR is evaluated in a zero-shot setting, further validating SKR's performance advantage. 


\subsection{Impact of Noise and Misleading Information}
\label{sec:misleading}
To further validate SKR's ability to eliminate noisy or misleading information, we replaced retrieved knowledge with two types of interference data: 1) Bad retrieval, where the order of the top-30 retrieved knowledge pieces is reversed to simulate scenarios where the retrieved knowledge is misleading. This type of data may still contain helpful information, but it also includes a significant amount of misleading information; 2) Random, where all retrieved knowledge is replaced with random samples from the entire corpus, to simulate a scenario where the retrieved information is irrelevant.
The experimental results are presented in Table~\ref{tab:noise}. 
We observed that both types of interference data negatively impacted the model. Taking Llama-2-7B as an example, the "Bad" type caused a decline of -9.53 and -9.71 in EM on NQ and WQ respectively, compared to not using retrieval data, while the "Random" type led to a decline of -9.3 and -7.17 in EM on NQ and WQ respectively. We then observed that SKR significantly mitigated the negative impact caused by severe interference in the retrieved data. Taking Llama-2-7B as an example, SKR under the "Bad" setting brought about EM improvements of +10.03 and +13.65 on NQ and WQ respectively. Under the "Random" setting, it resulted in EM improvements of +9.24 and +5.84 on NQ and WQ respectively.
These experimental results fully demonstrate that \textit{SKR can effectively eliminate noise and misleading information}.

\subsection{Effects on Compression Rate}
To verify SKR's capability to compress retrieved knowledge through rewriting, and to analyze the impact of supportiveness-based rewrite data
generation and supportiveness-based alignment from the perspective of compression rate, we conducted the experiments presented in Table~\ref{tab:comp}.

In this experiment, we used four different settings: 1) SKR w/o SRDG \& SA, which refers to training directly using the draft dataset generated by GPT-4; 2) SKR w/o SA, which means training using only the supportiveness-based rewrite data
generation; 3) SKR w/o SRDG, which implies training using only the supportiveness-based alignment; 4) SKR, which represents our complete method. We first observed that the full SKR achieved remarkable compression rates of 7.57x and 7.60x on the NQ and TQA datasets, respectively. Furthermore, we noted that supportiveness-based rewrite data
generation significantly positively impacts the compression rate (compression rate of 6.27x under the SKR w/o SRDG \& SA setting, and a rate of 7.49x under the SKR w/o SA setting). 
This can be attributed to the capability of recognizing irrelevant knowledge that SRDG brings to SKR.

\begin{table}[h]
    \centering
    \begin{minipage}{0.5\textwidth}
        \centering
        \small
        \begin{tabular}{@{}lllll@{}}
        \toprule
        \multicolumn{1}{c}{\multirow{2}{*}{Rewriter}} & \multicolumn{2}{c}{NQ}             & \multicolumn{2}{c}{WQ}             \\ \cmidrule(l){2-5} 
        \multicolumn{1}{c}{}                          & \multicolumn{1}{c}{length} & comp. & \multicolumn{1}{c}{length} & comp. \\ \midrule
        -                                             & 815.4                      & 1x    & 842.7                      & 1x    \\
        GPT-4                                         & 140.9                      & 5.79x & 161.1                      & 5.23x \\
        w/o SRDG \& SA                                & 130.1                      & 6.27x & 130.9                      & 6.44x \\
        w/o SA                                        & 108.8                      & 7.49x & 106.8                      & 7.89x \\
        w/o SRDG                                      & 121.4                      & 6.72x & 124.4                      & 6.77x \\
        SKR                                           & 107.7                      & 7.57x & 110.9                      & 7.60x \\ \bottomrule
        \end{tabular}
        \caption{\label{tab:comp}Effects on Compression Rate. We used Llama-2-7B as the evaluation model. ``length'' denotes the average length of the token sequences, and ``comp.'' represents the compression rate.}
        \vspace{6pt}
        
    \end{minipage}\hfill
    \begin{minipage}{0.45\textwidth}
        \centering
        \small
        
        \begin{tabular}{@{}llll@{}}
        \toprule
        \multicolumn{1}{c}{\multirow{2}{*}{Model}} & \multicolumn{1}{c}{\multirow{2}{*}{Rewriter}} & \multicolumn{2}{c}{NQ}                                 \\ \cmidrule(l){3-4} 
        \multicolumn{1}{c}{}                       & \multicolumn{1}{c}{}                          & \multicolumn{1}{c}{EM} & \multicolumn{1}{c}{Irr. rate} \\ \midrule
        \multirow{3}{*}{Llama-2-7B}                & GPT-4                                         & 28.2                   & 0\%                           \\
                                                   & GPT-4$\dagger$                                    & 29.82                  & 38.9\%                        \\
                                                   & SKR                                           & 31.22                  & 7.1\%                         \\ \bottomrule
        \end{tabular}
        \caption{\label{tab:irr}
Analysis of Recognizing Irrelevant Knowledge. ``Irr. rate'' denotes the rate at which the rewriter output ``irrelevant'' and ``GPT-4$\dagger$'' denotes using GPT-4 as the rewriter with modified prompt described in section~\ref{sec:4.5}.}
        \vspace{6pt}
        
    \end{minipage}
\end{table}

\subsection{Analysis of Recognizing Irrelevant Knowledge}
\label{sec:4.5}
By employing the supportiveness-based rewrite data
generation described in section~\ref{sec:2.2}, SKR has gained the capability to recognize irrelevant knowledge. Existing general-purpose LLMs like GPT-4 can also discern irrelevant knowledge through instruction prompts. To analyze and compare these two modes, we have constructed the evaluation shown in Table~\ref{tab:irr}.
Specifically, we first modified GPT-4's prompt to enable it to independently assess whether the retrieved knowledge contains information that aids in answering the query and to output "irrelevant" if it determines the information is not helpful. Subsequently, we recorded the rate at which different rewriter output "irrelevant", denoted as "Irr. rate," as shown in Table~\ref{tab:irr}. The modified prompt for GPT-4 is shown in Appendix A.



It can be observed that allowing GPT-4 to independently assess whether the retrieved knowledge contains information that aids in answering the query does not help improve the final performance. Additionally, it is noted that GPT-4 is quite stringent in identifying whether retrieved knowledge contains irrelevant information, with its Irr. rate significantly higher than SKR's (38.9\% vs. 7.1\%). We believe that GPT-4 is overly strict in its judgments, to the extent that it overlooks some knowledge that, while not overtly obvious, could provide supplementary lateral information to LLMs. Therefore, we utilize the proposed supportiveness-based rewrite data generation method to enable SKR to recognize irrelevant information, rather than relying on GPT-4 as the annotator for this task.

\subsection{Ablation Studies}
\label{sec:4.2}
To thoroughly analyze the contributions of supportiveness-based rewrite data
generation and supportiveness-based alignment to the performance of SKR, we conducted ablation experiments on the aforementioned two modules. 
The results of these experiments are presented in Table~\ref{tab:ablation1}. All settings uniformly utilize the top-5 knowledge chunks as retrieval data. ``SRDG'' denotes supportiveness-based rewrite data
generation, and ``SA'' denotes supportiveness-based alignment.

We observed that even without utilizing the two proposed supportiveness-based methods, SKR's rewriting performance is commendable. However, compared to GPT-4, there is a decrease in performance (SKR results in an average decrease of -0.79 in the EM metric for Llama-2-7B, across two tasks).
Another observation is that SRDG and SA individually contributed to performance improvements in SKR, ultimately surpassing the performance of GPT-4. SKR achieved an average increase of +2.00 in the EM metric for Llama-2-7B, across two tasks.

\begin{table}[th]
    \centering
    \begin{minipage}{0.45\textwidth}
        \centering
        \small
        \begin{tabular}{@{}llll@{}}
        \toprule
        \multicolumn{1}{c}{Rewriter} & \multicolumn{1}{c}{NQ} & \multicolumn{1}{c}{WQ} & \multicolumn{1}{c}{Avg.} \\ \midrule
        -                            & 17.5                   & 14.91                  & 16.21                    \\
        GPT-4                        & 29.82                  & 22.83                  & 26.33                    \\
        SKR w/o SRDG \& SA           & 29.33                  & 21.75                  & 25.54                    \\
        SKR w/o SA                   & 30.01                  & 22.61                  & 26.31                    \\
        SKR w/o SRDG                 & 30.48                  & 24.7                   & 27.59                    \\
        SKR                          & \textbf{31.42}         & \textbf{25.24}         & \textbf{28.33}           \\ \bottomrule
        \end{tabular}
        \caption{\label{tab:ablation1}Ablation with Llama-2-7B.}
    \end{minipage}\hfill
\end{table}

\section{Related Works}
Existing RALM research can be broadly classified into two categories. The first one primarily focuses on enhancing LLMs via pre-training or fine-tuning, aiming to improve their ability to effectively harness the retrieved knowledge~\cite{Borgeaud2021ImprovingLM, Guu2020REALMRL, Lin2023RADITRD, yu2023chainofnote}. Another line of RALM efforts explores directions beyond training LLMs, encompassing but not limited to 1) merging multiple generations based on diverse knowledge-enhanced queries~\cite{Shi2023REPLUGRB, Lin2023RADITRD}, 2) re-ranking candidate knowledge chunk~\cite{Ram2023InContextRL, jiang-etal-2023-llmlingua}, or 3) improving retrievers by training them with specific language tasks \cite{Shi2023REPLUGRB, Izacard2022FewshotLW}.
RALM can address the issues faced by large language models, such as the difficulty in timely updating information and the unreliable memory of long-tail knowledge~\cite{Wang2023ShallWP, Huang2022AreLP, Carlini2020ExtractingTD, Khandelwal2020Generalization, guu2020realm, ram2023incontext, izacard_few-shot_2022, asai2023selfrag, shi2023replug, khattab2022demonstrate, zhang-etal-2022-iclr,huang-etal-2022-nlpcc}.

However, the presence of noise or misleading information in the retrieved content can impair the performance of RALM and the extensive context information also introduces additional computational overhead~\cite{Wang2023LearningTF, Chen2023WalkingDT, Xu2023RetrievalML}. This has inspired research into compressing or rewriting the retrieved content.
RECOMP~\cite{Xu2023RECOMPIR} trains an extractive compressor based on the similarity between queries and sentences and uses GPT-3.5 as a teacher model to train an abstractive rewriter. 
CoN~\cite{yu2023chainofnote} employs GPT-3.5 as a teacher model, targeting answer prediction and context summarization simultaneously.
Most recently, Filco~\cite{Wang2023LearningTF} identifies the importance of sentences within the context by integrating multiple strategies. \citet{jiang2024longllmlinguaacceleratingenhancingllms} and \citet{jin2024biderbridgingknowledgeinconsistency} have sought to evaluate the quality of retrieved knowledge by examining the changes in perplexity (or probability) in the outputs of LLMs caused by the retrieved knowledge, thereby providing supervision signals for model training.

\section{Conclusion}
In this paper, we introduced the innovative concept of "supportiveness",  providing a simple yet powerful approach to evaluate how effectively a knowledge piece facilitates language tasks. 
Based on the supportiveness, we have developed the supportiveness-based rewrite data generation and supportiveness-guided alignment methods, harnessing them to train SKR, an effective knowledge rewriter.
Our comprehensive experiments across six datasets and four LLMs validate SKR's effectiveness and generalizability. With only 7B parameters, SKR demonstrates superior knowledge rewriting capabilities compared to GPT-4.

\bibliography{aaai25}
\newpage
\mbox{}
\newpage

\section*{Appendix}
\subsection{A. Prompts}

\begin{figure}[h]
\centering
\includegraphics[width=0.63\linewidth]{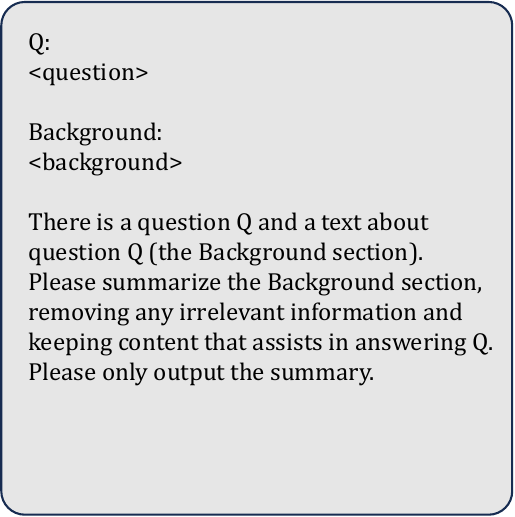}
\caption{Prompt for GPT-4.}
\label{fig:4}
\end{figure}

\begin{figure}[h]
\centering
\includegraphics[width=0.63\linewidth]{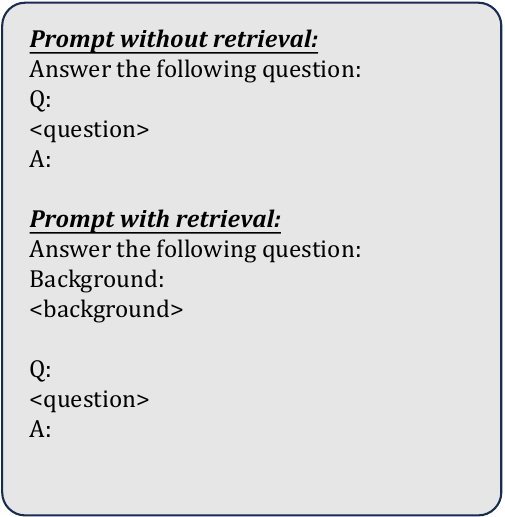}
\caption{Prompts for calculating supportiveness.}
\label{fig:5}
\end{figure}

\begin{figure}[h]
\centering
\includegraphics[width=0.63\linewidth]{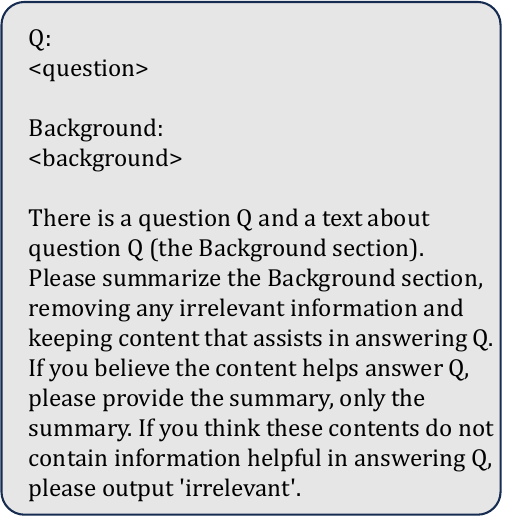}
\caption{Modified prompt for GPT-4.}
\label{fig:6}
\end{figure}

This section introduces the prompts mentioned in this paper. Figures \ref{fig:4}, \ref{fig:5}, and \ref{fig:6} illustrate the prompt for GPT-4, prompts for calculating supportiveness, and the modified prompt for GPT-4, which are discussed in the section Analysis of Recognizing Irrelevant Knowledge.

\subsection{B. Datasets}
We utilized three tasks encompassing a total of 10 datasets to train SKR. ``\#Size'' denotes the number of samples in the original training set of the dataset, while ``\#Train'' represents the number of training set samples used for training SKR after random sampling. 
The detailed statistics of these datasets can be found in Table 1.

\begin{table*}[ht]
\small
\centering
\setlength{\tabcolsep}{3.5mm}{\begin{tabular}{@{}llll@{}}
\toprule
Task                                   & Dataset name                                                                          & \#Size  & \#Train \\ \midrule
\multirow{4}{*}{Open-Domain QA}        & CommonsenseQA~\cite{Talmor2019CommonsenseQAAQ}                  & 9,741   & 9,741   \\
                                       & Yahoo! Answers QA                                                                     & 87,362  & 10,000  \\
                                       & Wiki Question Answering~\cite{yang-etal-2015-wikiqa}            & 20,360  & 10,000  \\
                                       & FreebaseQA~\cite{jiang-etal-2019-freebaseqa}                    & 20,358  & 10,000  \\ \midrule
\multirow{4}{*}{Reading Comprehension} & Conversational Question Answering~\cite{Reddy2018CoQAAC}        & 108,647 & 10,000  \\
                                       & Discrete Reasoning Over Paragraphs~\cite{dua-etal-2019-drop}    & 77,400  & 10,000  \\
                                       & QuaRel~\cite{Tafjord2018QuaRelAD}                               & 1,941   & 1,941   \\
                                       & SQuAD v2~\cite{rajpurkar-etal-2018-know}                        & 130,319 & 10,000  \\ \midrule
\multirow{2}{*}{Reasoning}             & Grade School Math 8K~\cite{Cobbe2021TrainingVT}                 & 7,473   & 7,473   \\
                                       & Explanations for CommonsenseQ~\cite{Aggarwal2021ExplanationsFC} & 7,598   & 7,598   \\ \bottomrule
\end{tabular}}
\caption{\label{t_data}
Datasets used for training SKR.}
\end{table*}

\subsection{C. Supplement to Main Results}
Due to space constraints, we have placed the experimental results using Llama-2-13B and Mistral-7B as the base models in Appendix C, Table 2. All settings are consistent with those described in the main results section of the paper.
\begin{table*}[ht]
\centering
\small
\begin{tabular}{@{}lllllllllll@{}}
\toprule
Model                        & Rewriter       & Retrieval & NQ             & TQA            & HoPo           & MMLU           & zsRE           & WQ             & Avg.           & $\Delta_{EM}$ \\ \midrule
\multirow{9}{*}{Mistral-7B}  & -              & -         & 25.73          & 59.16          & 21.28          & 58.92          & 19.66          & 21.8           & 34.43          & -             \\
                             & -              & Top-1     & 28.25          & 62.54          & 30.18          & 59.41          & 54.43          & 21.16          & 42.66          & -             \\
                             & -              & Top-5     & 26.62          & 60.42          & 27.17          & 60.91          & 30.61          & 17.32          & 37.18          & -             \\ \cmidrule(l){2-11} 
                             & GPT-3.5        & Top-5     & 28.99          & 61.06          & 30.67          & 61.23          & 53.27          & 20.32          & 42.59          & 5.41          \\
                             & GPT-4          & Top-5     & 32.04          & 62.8           & \textbf{33.39} & 61.88          & 56.46          & 20.08          & 44.44          & 7.27          \\
                             & RECOMP         & Top-5     & 28.02          & 60.99          & 30.63          & -              & -              & -              & -              & -             \\
                             & FILCO$\dagger$ & Top-5     & 31.01          & 62.67          &                & -              & 59.24          & 19.42          & -              & -             \\
                             & LongLLMLingua  & Top-5     & 27.74          & 61.44          & 30.1           & 60.47          & 59.11          & 20.19          & 43.18          &               \\
                             & SKR-7B         & Top-5     & \textbf{33.25} & \textbf{64.92} & 32.43          & \textbf{61.99} & \textbf{60.78} & \textbf{24.89} & \textbf{46.38} & 9.20          \\ \midrule
\multirow{9}{*}{Llama-2-13B} & -              & -         & 26.5           & 61.02          & 21.63          & 49.15          & 23.2           & 24.05          & 34.26          & -             \\
                             & -              & Top-1     & 31.83          & 64.32          & 31.35          & 50.92          & 63.21          & 25.14          & 44.46          & -             \\
                             & -              & Top-5     & 26.12          & 62.32          & 29.51          & 52.32          & 61.98          & 19.63          & 41.98          & -             \\ \cmidrule(l){2-11} 
                             & GPT-3.5        & Top-5     & 35.23          & 66.12          & 29.44          & 52.91          & 64.28          & 27.22          & 45.87          & 3.89          \\
                             & GPT-4          & Top-5     & 36.66          & 67.19          & 34.3           & 52.94          & 65.64          & 28.73          & 47.58          & 5.6           \\
                             & RECOMP         & Top-5     & 34.17          & 66.08          & 29.25          & -              & -              & -              & -              & -             \\
                             & FILCO$\dagger$ & Top-5     & 35.42          & 65.43          & 32.03          & -              & 64.17          & 27.3           & -              & -             \\
                             & LongLLMLingua  & Top-5     & 34.19          & 64.42          & 28.85          & 50.76          & 64.12          & 27.19          & 44.92          &               \\
                             & SKR-7B         & Top-5     & \textbf{38.11} & \textbf{67.21} & \textbf{35.07} & \textbf{54.29} & \textbf{65.98} & \textbf{29.71} & \textbf{48.40} & 6.41          \\ \bottomrule
\end{tabular}
\caption{\label{tab:Supplement}Supplement to Main Results. ``$\dagger$'' denotes our implementations.}
\end{table*}

\subsection{D. Impact of Retriever Selection}
\label{Appendix:D}
Furthermore, to investigate the generalizability of SKR, we also employed another popular method, Contriever~\cite{izacard2022unsupervised}, as the retriever in our evaluation. The usage of Contriever slightly differs from DRAGON+. Specifically, Contriever uses a shared encoder $\mathbf{E}$ to encode both query and knowledge chunks, the similarity score is calculated as:
\begin{equation}
    s(q,c) = \mathbf{E}(c) \cdot \mathbf{E}(q).
\end{equation}
The experimental results are shown in Table~\ref{tab:contriever}.
We observed that due to Contriever's weaker performance compared to DRAGON+, the average performance of Llama-2-7B with top-5 retrieved knowledge dropped by -1.56 EM on the two datasets. Both SKR and GPT-4 can successfully rewrite knowledge under different retriever settings, and SKR still maintains its performance lead over GPT-4 (+1.93 EM). This confirms the robustness of SKR to different retrievers.

\begin{table}[ht]
\centering
\begin{tabular}{@{}lllll@{}}
\toprule
Retriever                   & Rewriter & NQ    & WQ    & Avg.  \\ \midrule
\multirow{3}{*}{DRAGON+}    & -        & 17.5  & 14.91 & 16.21 \\
                            & GPT-4    & 29.82 & 22.83 & 26.33 \\
                            & SKR      & 31.42 & 25.24 & 28.33 \\ \midrule
\multirow{3}{*}{Contriever} & -        & 14.8  & 14.13 & 14.47 \\
                            & GPT-4    & 27.12 & 22.56 & 24.84 \\
                            & SKR      & 28.74 & 24.79 & 26.77 \\ \bottomrule
\end{tabular}
\caption{
\label{tab:contriever}
Impact of Retriever Selection. ``Retriever'' denotes the use of different models as the retriever. ``Avg.'' denotes the average performance across the two datasets. In this experiment, we use the top-5 knowledge pieces as retrieval data and evaluate EM using Llama-2-7B.}
\end{table}

\subsection{E. Effects on Feedback Providers}
\label{Appendix:E}
Table~\ref{tab:10} resents the performance of SKR across four tasks when utilizing different feedback providers. In this experiment, we consistently employ the Llama-2-7B model for answer prediction, with all other settings aligned with those described in the section Experiment Results. Our observations indicate that the variations in performance due to different feedback providers are negligible, affirming the robustness of the proposed supportiveness evaluation methodology.

\begin{table}[ht]
\centering
\small
\begin{tabular}{@{}llllll@{}}
\toprule
\multirow{2}{*}{Feedback   Provider} & \multirow{2}{*}{NQ} & \multirow{2}{*}{TQA} & \multirow{2}{*}{HoPo} & \multirow{2}{*}{WQ} & \multirow{2}{*}{Avg.} \\
                                     &                     &                      &                       &                     &                       \\ \midrule
Mistral-7B                           & 31.42               & 63.62                & 30.98                 & 25.24               & 37.82                 \\
Llama-2-7B                           & 31.44               & 63.58                & 30.86                 & 25.11               & 37.75                 \\
Llama-2-13B                          & 31.37               & 63.44                & 31.04                 & 25.17               & 37.76                 \\ \bottomrule
\end{tabular}
\caption{\label{tab:10}Effects on Feedback Providers.}
\end{table}

\newpage
\end{document}